\documentclass{article} 
\usepackage{nips15submit_e,times}
\usepackage{hyperref}
\usepackage{url}

\usepackage{graphicx}
\usepackage{amsmath}
\usepackage{amssymb}

\usepackage{booktabs}
\usepackage{multirow}
\usepackage{tabularx}

\usepackage{algorithm}
\usepackage{algorithmic}

\usepackage{color}

\hypersetup{
  pdfinfo={
    Title={Multimodal Transfer Deep Learning with Applications in Audio-Visual Recognition},
    Author={Seungwhan Moon},
    Keywords={transfer learning, multimodal deep learning, audio-visual recognition}
  }
}

\pdfoutput=1

\title{Multimodal Transfer Deep Learning with Applications in Audio-Visual Recognition}

\author{
Seungwhan Moon$^*$, Suyoun Kim$^\dagger$, Haohan Wang$^\ddagger$ \\
$^*$Language Technologies Institute, $^\dagger$Electrical and Computer Engineering, $^\ddagger$Machine Learning Department\\
Carnegie Mellon University\\
\texttt{\{seungwhm@cs\hspace{1pt}|\hspace{1pt}suyoung1@ece\hspace{1pt}|\hspace{1pt}haohanw@cs\}.cmu.edu} \\
}

\nipsfinalcopy 

\begin{document}
\maketitle

\begin{abstract}
We propose a transfer deep learning (TDL) framework that can transfer the knowledge obtained from a single-modal neural network to a network with a different modality. Specifically, we show that we can leverage speech data to fine-tune the network trained for video recognition, given an initial set of audio-video parallel dataset within the same semantics. Our approach first learns the analogy-preserving embeddings between the abstract representations learned from intermediate layers of each network, allowing for semantics-level transfer between the source and target modalities. We then apply our neural network operation that fine-tunes the target network with the additional knowledge transferred from the source network, while keeping the topology of the target network unchanged. While we present an audio-visual recognition task as an application of our approach, our framework is flexible and thus can work with any multimodal dataset, or with any already-existing deep networks that share the common underlying semantics. In this work in progress report, we aim to provide comprehensive results of different configurations of the proposed approach on two widely used audio-visual datasets, and we discuss potential applications of the proposed approach.
\end{abstract}

\vspace{-7pt}
\section{Introduction}
\vspace{-10pt}

Multimodal deep networks have been recently proposed to leverage the features learned from multiple modalities to predict patterns of single or multiple modalities (\cite{Ngiam+11a, srivastava2012multimodal}). While the main focus of this line of work has been to construct a shared representation that best combines multiple modalities, most of the work assume the existence of the parallel multimodal dataset where the shared multimodal representation can be learned.

In reality, however, acquiring a parallel multimodal dataset is extremely resource consuming, and thus there is often an imbalance in the amount of the labeled data among different modalities. For example, while the labeled audio speech data is readily abundant, the labeled data for lip reading videos is much more scarce, consequently making the unparallel portion of the audio data obsolete for multimodal learning. In this paper, we address the data imbalance among different modalities via a transfer learning approach.

The transfer learning relaxes the imbalance of the available data in source tasks and target tasks via selective instance selections or feature transformation (\cite{Taylor+07, Pan+08}). 
However, while transfer learning works well when source tasks and target tasks are moderately in the same domain, direct knowledge transfer between different modalities is often intractable often due to their drastically different statistical properties in concept space (\cite{Navarretta+13}). 
In this regard, several approaches such as DCCA (\cite{DCCA}), DeViSE (\cite{DeViSE}), HHTL (\cite{HHTL}), and DCCAE (\cite{DCCAE}) have proposed to utilize deep neural networks to obtain abstract representation of data (\textit{e.g}. at the top-most layer of the network), thus allowing for more viable knowledge transfer. Most of these approaches aim at learning a single robust mapping between the two modalities, and then use the newly learned embeddings to learn a new model for a shared task or a target task. 

In this paper, we propose a new approach that can benefit the target task in the case of source-target data imbalance scenario. Specifically, we learn the semantic mappings between intermediate layers of each neural network, and leverage learned embeddings to fine-tune the target network. Our approach has several practical benefits in that we keep the topology of the target network unchanged, thus not requiring to build a separate model with a shared representation, etc. Instead, we directly modify the target network by tuning its hyper-parameters, and thus the same network can be used for the target task.

As an application of our framework, we choose to study the transfer between speech and lip-reading video data. However, our framework is unobtrusive and flexible, and thus can be used for transfer between any modalities with any two already-built deep networks, given the small initial parallel modality corpora that correspond to the same semantics.

\begin{figure}[t!]
  \centering
  \includegraphics[width=0.9\columnwidth]{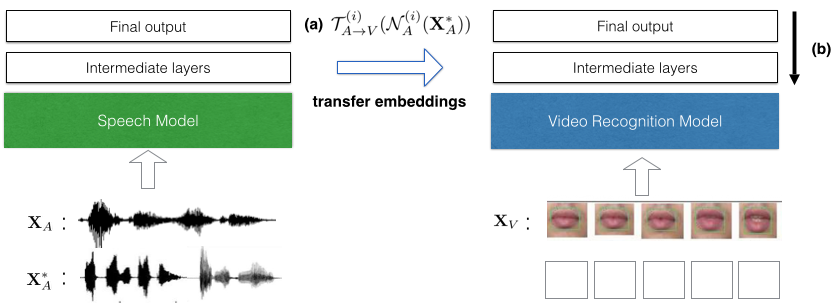}
  \caption{An illustration of our approach with an application in audio and lip-reading video recognition tasks. Notations are defined in Section \ref{sec:problem_formulation}. (a) We learn the embeddings between abstract representations of audio and video data using the parallel dataset ($\mathbf X_A$ and $\mathbf X_V$), from which we can transfer the new audio input $\mathbf X_A^*$, unforeseen in the label space of $\mathbf X_V$. (b) We can then fine-tune the video network $\mathcal N_V$ with the transferred audio data  $\mathcal T_{A \rightarrow V} ( \mathcal N_A^{(i)} ( \mathbf X_A^{*} ) ) $.} 
  \vspace{-10pt}
  \label{fig:overview}
\end{figure}

\vspace{-7pt}
\section{Problem Formulation}
\label{sec:problem_formulation}
\vspace{-10pt}

We formulate the proposed framework with application in the audio and video lip-reading multimodal learning settings. Figure \ref{fig:overview} illustrates the following formulation. We initially have two input data of different modalities: $\mathbf X_A = \{X_{A}^{1}, \cdots, X_{A}^{N} \}$ and $\mathbf X_V = \{X_{V}^{1}, \cdots, X_{V}^{N} \}$, which correspond to the parallel audio and video data, respectively. Both $\mathbf X_A$ and $\mathbf X_V$ map to the same ground-truth categorical labels $\mathbf Z = \{Z^{1}, \cdots, Z^{N} \}$. Each input audio instance $X_{A}^{n}$ and video instance $X_{V}^{n}$ lies in different concept spaces, thus $X_{A}^{n} \in \mathbb R^{p}$ and $X_{V}^{n} \in \mathbb R^{q}$. Two separate neural nets, $\mathcal N_A$ and $\mathcal N_V$, can be built from $\mathbf X_A$ and $\mathbf X_V$, respectively. We denote $\mathcal N_A: X_A \rightarrow Y$ and $\mathcal N_V: X_V \rightarrow Y$ where $Y$ is a predicted label as a final output of the neural networks, given some $X_A \in \mathbf X_A$ or $X_V \in \mathbf X_V$. 

Additionally, we also denote  $\mathcal N_A^{(i)}: X_A \rightarrow H_A^{(i)}$ and $\mathcal N_V^{(i)}: X_V \rightarrow  H_V^{(i)}$, where $H_A^{(i)} \in \mathbb R^{p_i}$ and $H_V^{(i)} \in \mathbb R^{q_i}$ are the output of the $i$-th layer of the two neural nets, respectively, $\forall i \in \{0,1,\cdots,l+1\}$. 
Note that $H^{(0)} = X$, $H^{(l+1)} = Y$, and $H^{(i+1)} := g(H^{(i)}, \mathcal W^{(i \rightarrow i+1)})$, where $\mathcal W^{(i \rightarrow i+1)}$ is the weight between $H^{(i)}$ and $H^{(i+1)}$ for $i \in \{0,\cdots,l\}$. $H_A$ and $H_V$ thus represent the input $X_A$ and $X_V$ at different abstraction levels. 

We then define a new set of labeled audio data, ($\mathbf X_A^* = \{{X_A^*}^1, \cdots, {X_A^*}^{M} \},\mathbf Z^* = \{{Z^*}^{1}, \cdots, {Z^*}^{M} \})$, which is unparallel and unforeseen in the video dataset in label space ($\mathbf X_V^*$ is assumed to be not available at training phase). Our purpose is then to learn a transfer function $\mathcal T_{A \rightarrow V}^{(i)} : H_A^{(i)} \rightarrow H_V^{(i)}$ $\forall i \in \{0,1,\cdots,m\}$ using $\mathbf X_A$ and $\mathbf X_V$, from which we can fine-tune $\mathcal N_V$ with $\mathcal T_{A \rightarrow V} (H_A^*)$,
where $H_{A}^{*} = \mathcal N_A (X_{A}^{*})$ for $X_A^* \in \mathbf X_A^*$  (Section \ref{sec:method}). 

\vspace{-7pt}
\section{Method}
\label{sec:method}
\vspace{-10pt}

We obtain abstract representations of the raw data using a standard deep belief network (DBN) with multiple RBM layers. \cite{wang2015survey}
Output values at each intermediate layer of a DBN thus give abstract feature representation of the input data, allowing for a more tractable knowledge transfer between modalities. In our experiment, we build two DBNs ($\mathcal N_A$ and $\mathcal N_V$) for audio and video data that have the same number of intermediate layers, and we learn inter-modal embeddings for each layer at the same depth as detailed in Section \ref{subsec:method_embeddings}. Using the learned mapping between the source and target modalities, we fine-tune the network with the transferred data as detailed in Section \ref{subsec:method_fine_tuning}.

\subsection{Learning the Embeddings}
\label{subsec:method_embeddings}
\vspace{-10pt}

We learn the embedding function $\mathcal T_{A \rightarrow V}$ which maps two concept spaces $H_A^{(i)} \in \mathbb R^{p_i}$ and $H_V^{(i)} \in \mathbb R^{q_i}$. While any embedding method can be applied, we consider the following three embedding methods from literature.

\textbf{Multivariate Support Vector Regression (SVR) Using Nonlinear Kernels}: we formulate the mapping between two concept spaces as a multivariate regression problem. Specifically, we use Support Vector Regression (SVR) methods with nonlinear kernels (\cite{Smola+04}), which can effectively learn the conditional expectation of the target space given the source space. For practical use, we kernelize the weights and add a soft-margin loss function for more flexible and high-dimensional mapping. 

\textbf{KNN-based Non-parametric Mapping}: 
we can also obtain mappings of new audio input by first finding the $K$-closest audio samples in the training set, and then returning the average of the values of the corresponding video samples. 

\textbf{Normalized Canonical Correlation Analysis (NCCA)}: given a set of $N$ audio and video pairs $\mathbf H_A \in \mathbb R^{N \times p'}$ and $\mathbf H_V \in \mathbb R^{N \times q'}$, NCCA obtains $\mathbf U \in \mathbb R^{p' \times c}$ and $\mathbf V \in \mathbb R^{q' \times c}$ which project audio and video into a common $c$-dimensional latent space by $\mathbf H_A \mathbf U$ and $\mathbf H_V \mathbf V$ (\cite{gunhee-cvpr-15-text2pic}). The objective can thus be formulated as:

{
\vspace{-12pt}
\begin{align}
\label{eq:cca}
\max_{\mathbf U, \mathbf V} & \mbox{ tr} (\mathbf U^T \hspace{-1pt} \mathbf H_A^T \hspace{-1pt}  \mathbf H_V \mathbf V)  \\
\mbox{ s.t. } &\mathbf U^T \hspace{-1pt}  \mathbf H_A^T \hspace{-1pt}  \mathbf H_A \mathbf U \hspace{-2pt}=\hspace{-2pt} \mathbf I, 
\mathbf V^T \hspace{-1pt}  \mathbf H_V^T \hspace{-1pt}  \mathbf H_V \mathbf V \hspace{-2pt}=\hspace{-2pt} \mathbf I. \nonumber
\end{align}
\vspace{-12pt}
}

\noindent which we can solve as a generalized eigenvalue problem.

Once $\mathbf U$ and $\mathbf V$ are obtained, we can formulate the mapping function as $\mathcal T_{A \rightarrow V} (\mathbf H_{A}) = \mathbf H_{A} \mathbf U \mathbf V^T$ to obtain the estimated transfer $\mathbf H_{V}$.

\subsection{Fine-tuning with the Transferred Data}
\label{subsec:method_fine_tuning}
\vspace{-10pt}

We propose an algorithm to fine-tune the network with the transferred data (\texttt{TDLFT}) as described in Algorithm \ref{alg:tdlft}. \texttt{TDLFT(i)} obtains a transfer of the new audio input at $i$-th layer, $H_V^{(i)} := \mathcal T_{A \rightarrow V}^{(i)}(\mathcal N_A^{(i)}(X_A^*))$, treats it as an input to $\mathcal N_V$ at $i$-th layer, and computes $H_V^{(j+1)} := g(H_V^{(j)}, \mathcal W_V^{(j \rightarrow j+1)})$ for $j \in \{i,i+1,\cdots,l\}$, where $g$ is an activation function.
Finally, we fine-tune $\mathcal W_V^{(j \rightarrow j+1)}$ for $j \in \{i, i+1, \cdots, l \}$ via a standard backpropagation algorithm.

Note that \texttt{TDLFT(i)} does not fine-tune $\mathcal W_V^{(j \rightarrow j+1)}$ for $j \in \{0, 1, \cdots, i-1 \}$. Therefore, only the top  $(l-i+1)$ layers are fine-tuned to the new stimuli, which could be a drawback of this algorithm. However, we argue that this issue can be naturally mitigated through some good choice of $i$, at which a typical error backpropagation technique reaches the inevitable \textit{vanishing errors} at bottom layers \cite{glorot2010understanding}. Assuming a perfect transfer function learned at this choice of $i$, we almost have the same effect as if we had a parallel video data $X_V^*$ available for training.

It is interesting to note the expected trade-off behavior for different choices of $i$. When $i$ is big ($i \simeq l$), we expect that the accuracy of transfer is more reliable (e.g. intuitively, $\mathcal T_{A \rightarrow V}^{(l+1)}$ should always be correct, because it is a mapping from and to the same label space), however we only get to fine-tune the top $(l-i+1)$ layer(s), leaving the rest of the bottom layers un-tuned. A smaller choice of $i$ mitigates this issue, however we typically suffer from an unreliable transfer for $i \simeq 0$ (e.g. $\mathcal T_{A \rightarrow V}^{(0)}$ is almost intractable, given drastically different concept spaces of the two input types).

\begin{algorithm}[tb]
\caption{Transfer Deep Learning Fine-Tuning (TDLFT)}
\begin{algorithmic}

\STATE {\bfseries Input}: $\mathcal N_A$ trained with $\mathbf X_A$, $\mathcal N_V$ trained with $\mathbf X_V$, an input parameter $i \in \{0,1,\cdots,l\}$, $\mathcal T_{A \rightarrow V}^{(j)} : \mathbf H_A^{(j)} \rightarrow \mathbf H_V^{(j)}$ learned for $j:=i$, a new unparallel data ($\mathbf X_{A}^{*}$, $\mathbf Y^{*}$).

\STATE {\bfseries Output:}  $\mathcal N_V$ fine-tuned with $\mathbf X_{A}^{*}$.

\STATE Obtain $\mathbf H_V^{(i)} := \mathcal T_{A \rightarrow V}^{(i)}(\mathcal N_A^{(i)}(X_A^*))$, and $\mathbf H_V^{(l+1)} := \mathbf Y^{*}$

\FOR { $j \in \{i,i+1,\cdots,l\}$ }
    \STATE $\mathbf H_V^{(j+1)} := g(\mathbf H_V^{(j)}, \mathcal W_V^{(j \rightarrow j+1)})$ 
\ENDFOR
\STATE Fine-tune $\mathcal W_V^{(j \rightarrow j+1)}$ for $j \in \{i,i+1,\cdots,l\}$ via a standard backpropagation algorithm.

\end{algorithmic}
\label{alg:tdlft}
\end{algorithm}


\vspace{-7pt}
\section{Empirical Evaluation}
\label{sec:experiment}
\vspace{-10pt}

\subsection{Dataset}
\label{subsec:experiment_dataset}
\vspace{-10pt}

We use two widely used audio-visual multimodal datasets for empirical evaluation, \texttt{AV-Letters} and \texttt{Stanford}. The \texttt{AV-Letters} dataset (\cite{Matthews+02}) consists of both audio and lip-reading video data of 10 speakers saying the letters A to Z (thus 26 labels), three times each. The dataset contains pre-extracted lip regions at $60\times80$ pixels. We represent each audio example as one feature vector by concatenating 24 contiguous audio frames, each with 26 mel-frequency cepstral coefficients (MFCC). 
Similarly, for each video example, we concatenate 12 contiguous video frames, each with $60\times80$ pixels. 
The \texttt{Stanford} dataset (\cite{Ngiam+11a}) consists of both audio and lip-reading video data of 23 speakers saying the digits 0 to 9, letters A to Z, and some selected sentences (total 49 labels). 

We divide each dataset by its label space to simulate the data imbalance scenario. For instance, in the \texttt{AV-Letters} dataset, we assume that only the data with labels from 1 to 20 (denoted $\mathbf X_V$) are available as a parallel corpus during training of video data, and the video data with labels ranging from 21 to 26 (denoted $\mathbf X_{V}^{*}$) are assumed to be completely unforeseen during the training phase. On the other hand, we assume that the audio data has access to the entire label space (labels 1 to 26) during training ($\mathbf X_{A} \cup \mathbf X_{A}^{*}$). This is analogous to many real-world situations where the training data for multi-modal learning is imbalanced for one of the modalities, thus the test data of the target modality is radically different from the training data. Table \ref{tab:dataset} summarizes the datasets configuration.

\begin{table}
    \vspace{-10pt}
    \begin{center}
    \caption{Overview of Datasets.}
    \label{tab:dataset}
    \begin{tabular}{rcccc}
    \toprule
    \   Dataset & Division & Labels & \# Attribtues & \# Instances  \\ 
    \midrule
    \  \multirow{4}{*}{\texttt{AV-Letters}} & $\mathbf X_{A}$  & 1-20 & \multirow{2}{*}{624} & 600 \\ 
    \   & $\mathbf X_{A}^{*}$ & 21-26 &  & 180 \\ 
    \cmidrule(r){2-5}
    \   & $\mathbf X_{V}$ & 1-20 & \multirow{2}{*}{57,600} & 600 \\ 
    \   & $\mathbf X_{V}^{*}$ & 21-26  & & 180 \\ 
    \midrule
    \  \multirow{4}{*}{\texttt{Stanford}} & $\mathbf X_{A}$  & 1-44 & \multirow{2}{*}{1,573} & 2,064 \\ 
    \   & $\mathbf X_{A}^{*}$ & 45-49 &  & 504 \\ 
    \cmidrule(r){2-5}
    \   & $\mathbf X_{V}$ & 1-44 & \multirow{2}{*}{19,481} & 2,064 \\ 
    \   & $\mathbf X_{V}^{*}$ & 45-49  & & 504 \\ 
    \bottomrule
    \end{tabular}
    \end{center}
    \vspace{-15pt}
\end{table}

\subsection{Task}
\label{subsec:experiment_task}
\vspace{-10pt}

To evaluate how the transferred knowledge from the source modality improves classification performance of the target modality, we perform comparison studies on the following task with the \texttt{AV-Letters} and \texttt{Stanford} dataset (in Section \ref{subsec:experiment_dataset}). The task is to build a classifier that categorizes each lip-reading video into a label (1-26 for \texttt{AV-Letters}, and 1-49 for \texttt{Stanford}). We assume that only $\mathbf X_V$ and $\mathbf X_A$ are available as a parallel corpus during initial training.

Assuming that the audio data has extra labeled data ($\mathbf X_{A}^{*}$) unparallel in the video training set, we fine-tune the video model ($\mathcal N_V$) with the extra transferred audio data as described in Section \ref{subsec:method_fine_tuning}. The classification task is performed in a cross validation way on the data spanning the entire labels ($\mathbf X_{V} \cup \mathbf X_{V}^{*}$). We then compare the result of our transfer deep learning approach (\texttt{TDL}) against the \texttt{unimodal} baseline where the model is trained only with the available video data ($\mathbf X_V$).

\subsection{Results}
\vspace{-10pt}

In this section, we provide comprehensive empirical results on the task described in Section \ref{subsec:experiment_task} with different combinations of embedding transfer methods and fine-tuning methods (in Section \ref{sec:method}). 

\begin{table}
    \vspace{-10pt}
    \begin{center}
    \caption{5-fold lip-reading video classification accuracy on the \texttt{AV-Letters} dataset (26 labels)}
    \label{tab:av_letters_result}    
    \begin{tabular}{rccc}
    \toprule
    \        & \texttt{Unimodal} & \textbf{\texttt{TDL}} & \texttt{Oracle}  \\ 
    \cmidrule(r){2-4}
    \ Train: &  $\mathbf X_V$       & $\mathbf X_V \cup  \underline{\mathcal T_{A \rightarrow V} (\mathbf H_{A}^{*})}$ & $\mathbf X_V \cup  \underline{\mathbf X_{V}^{*}}$  \\ 
    \ Test:  & $\mathbf X_{V} \cup \mathbf X_{V}^{*}$ & $\mathbf X_{V} \cup \mathbf X_{V}^{*}$ & $\mathbf X_{V} \cup \mathbf X_{V}^{*}$ \\ 
    \midrule
    \ \textbf{\texttt{KNN + TDLFT(3)}} & \multirow{3}{*}{ 51.1\%} &  \textbf{55.3}\% & \multirow{3}{*}{ 61.7\%} \\ 
    \ \textbf{\texttt{NCCA + TDLFT(3)}} & &  \textbf{53.1}\% &  \\ 
    \ \textbf{\texttt{SVR + TDLFT(3)}} &  &  \textbf{54.8}\% &  \\
    \midrule
    \ \texttt{KNN + TDLFT(0)} & \multirow{3}{*}{ 51.1\%} &  34.4\% & \multirow{3}{*}{ 66.8\%}  \\
    \ \texttt{NCCA + TDLFT(0)} &  &  32.1\% &  \\    
    \ \texttt{SVR + TDLFT(0)} &  & 48.3\% &   \\    
    \bottomrule
    \end{tabular}
    \end{center}
    \vspace{-15pt}
\end{table}

\begin{table}
    \vspace{-5pt}
    \begin{center}
    \caption{5-fold lip-reading video classification accuracy on the \texttt{Stanford} dataset (49 labels)}
    \label{tab:stanford_result}    
    \begin{tabular}{rccc}
    \toprule
    \        & \texttt{Unimodal} & \textbf{\texttt{TDL}} & \texttt{Oracle}  \\ 
    \cmidrule(r){2-4}
    \ Train: &  $\mathbf X_V$       & $\mathbf X_V \cup  \underline{\mathcal T_{A \rightarrow V} (\mathbf H_{A}^{*})}$ & $\mathbf X_V \cup  \underline{\mathbf X_{V}^{*}}$  \\ 
    \ Test:  & $\mathbf X_{V} \cup \mathbf X_{V}^{*}$ & $\mathbf X_{V} \cup \mathbf X_{V}^{*}$ & $\mathbf X_{V} \cup \mathbf X_{V}^{*}$ \\ 
    \midrule
    \ \textbf{\texttt{KNN + TDLFT(3)}} & \multirow{3}{*}{ 54.9\%} &  \textbf{61.3}\% & \multirow{3}{*}{ 68.2\%} \\ 
    \ \textbf{\texttt{NCCA + TDLFT(3)}} & &  \textbf{58.2}\% &  \\ 
    \ \textbf{\texttt{SVR + TDLFT(3)}} &  &  \textbf{56.8}\% &  \\
    \midrule
    \ \texttt{KNN + TDLFT(0)} & \multirow{3}{*}{ 54.9\%} &  49.8\% & \multirow{3}{*}{ 73.2\%}  \\
    \ \texttt{NCCA + TDLFT(0)} &  &  52.4\% &  \\    
    \ \texttt{SVR + TDLFT(0)} &  & 45.3\% &   \\    
    \bottomrule
    \end{tabular}
    \end{center}
    \vspace{-15pt}
\end{table}

Tables \ref{tab:av_letters_result} and \ref{tab:stanford_result} shows the comparison of 5-fold cross validation performance between the \texttt{unimodal} (video) model and the transfer deep learning model (\texttt{TDL}) where the network is additionally fine-tuned with the transferred audio data.
\texttt{KNN}, \texttt{NCCA}, and \texttt{SVR} refer to the embedding methods that were applied to transfer the audio representations, as detailed in Section \ref{subsec:method_embeddings}. \texttt{TDLFT(i)} refers to the fine-tuning method applied starting at the $i$-th layer of the network (Section \ref{subsec:method_fine_tuning}). 
The \texttt{Oracle} baseline shows the semi-oracle bound of \texttt{TDL} which can be achieved if the perfect knowledge transfer $\hat{\mathcal T_{A \rightarrow V}}$ is obtainable. The underlined portion on the train sets denote the data that was used to further fine-tune the network via \texttt{TDLFT}. Bold denotes significant improvement of \texttt{TDL} over the baseline. The chance performance for \texttt{AV-Letters} and \texttt{Stanford} is 3.85\% (=1/26) and 2.04\% (=1/49), respectively.

\texttt{TDL} with \texttt{TDLFT(3)} outperforms \texttt{unimodal} for both datasets when audio data (source domain) was transferred with any of the proposed embedding methods, showing that the transferred modality enhances the performance of the target modality classification. This is promising because it indicates that we can improve the classification performance in the previously unknown classes of the target task without training samples for those classes. Specifically, our results show that the KNN-based transfer method (\texttt{KNN}) generally yields the best embeddings, followed by \texttt{NCCA} and \texttt{SVR}. 

Note that the \texttt{TDL} performance at \texttt{TDLFT(0)} (where the transfer between modalities was at the raw feature representation level) does not show a significant improvement over the \texttt{Unimodal} baseline. This is because learning embeddings at the input level between the target and the source modalities is often intractable, leading to a poor transfer performance. The network is then fine-tuned with noised transferred data, thus negatively affecting the classification performance. Note that this approach is similar to the traditional feature-transformation-based transfer learning methods.

All of the \texttt{TDL} performances are upper-bounded by the \texttt{Oracle} performance which simulates a perfect transfer between modalities. This result indicates that we can improve the transfer deep learning performance with a better transfer embedding method between modalities. Note also that the \texttt{Oracle} performance with \texttt{TDLFT(0)} has the best overall performance. While the transfer of embeddings at the raw input level is intractable in reality, \texttt{TDLFT(0)} fully fine-tunes every layer in the network, which would thus improve the performance the most if embeddings are reliable.

\vspace{-7pt}
\section{Conclusions}
\label{sec:conclusion}
\vspace{-10pt}

We proposed a framework for performing transfer learning on neural networks (TDL) under multimodal learning settings. We proposed several embedding methods for transferring knowledge between the target and source modalities, and presented our results on two audio-visual datasets. Our results show that the transferred modality of an abstract representation obtained from intermediate layers of the source network can be effectively utilized to further fine-tune the target network. Specifically, our results indicate that our approach is especially applicable when the data in the target modality is much more scarce (\textit{i.e.} in label space) than in the source modality.

\textbf{Future Work}: we note that the proposed framework can be extended to the idea of reconstructing the modality given the transferred modality via top-down inference \cite{Lee+09}. This has many potential applications, for instance in automatically generating lip-motion videos given any audio input using the transferred audio input, which is an improvement over the conventional lip-motion generator that is heavily rule-based or human-engineered. We plan on addressing this problem in the future work.

\vspace{-8pt}
\bibliography{bibliography}
\bibliographystyle{ieeetr}
\end{document}